\documentclass[10pt, a4paper]{article}
\usepackage{lrec2022} 
\usepackage{multibib}
\newcites{languageresource}{Language Resources}
\usepackage{graphicx}
\usepackage{tabularx}
\usepackage{soul}
\usepackage{adjustbox}
\usepackage{titlesec}
\titleformat{\section}{\normalfont\large\bfseries\center}{\thesection.}{1em}{}
\titleformat{\subsection}{\normalfont\SmallTitleFont\bfseries\raggedright}{\thesubsection.}{1em}{}
\titleformat{\subsubsection}{\normalfont\normalsize\bfseries\raggedright}{\thesubsubsection.}{1em}{}
\renewcommand\thesection{\arabic{section}}
\renewcommand\thesubsection{\thesection.\arabic{subsection}}
\renewcommand\thesubsubsection{\thesubsection.\arabic{subsubsection}}

\usepackage{epstopdf}
\usepackage[utf8]{inputenc}

\usepackage{hyperref}
\usepackage{url}
\usepackage{xstring}

\usepackage{color}
\newcommand{\specialcell}[2][c]{%
  \begin{tabular}[#1]{@{}l@{}}#2\end{tabular}}

\usepackage{authblk}

\title{The Project Dialogism Novel Corpus:\\ A Dataset for Quotation Attribution in Literary Texts}
\name{Krishnapriya Vishnubhotla$^{\ast}$$^{\dagger}$, Adam Hammond$^{\ddagger}$, Graeme Hirst$^{\ast}$}

\address{$^{\ast}$Department of Computer Science, University of Toronto \\ 
            $^{\ddagger}$Department of English, University of Toronto \\
        $^{\dagger}$Vector Institute for Artificial Intelligence \\
          \texttt{vkpriya@cs.toronto.edu, adam.hammond@utoronto.ca, gh@cs.toronto.edu}}

\abstract{
We present the Project Dialogism Novel Corpus, or PDNC, an annotated dataset of quotations for English literary texts. PDNC contains annotations for 35,978 quotations across 22 full-length novels, and is by an order of magnitude the largest corpus of its kind. Each quotation is annotated for the speaker, addressees, type of quotation, referring expression, and character mentions within the quotation text. The annotated attributes allow for a comprehensive evaluation of models of quotation attribution and coreference for literary texts.
 \\ \newline  }

\begin{document}

\maketitleabstract

\section{Introduction}
Computational analysis of literary texts looks into Natural Language Processing (NLP) techniques to model aspects of narrative, events, and characters \cite{elsner2012character,bamman2014bayesian,vishnubhotla2019fictional}. Past work in this area has focused mainly on analysing works of fiction, drawn from open-source platforms like Project Gutenberg\footnote{https://www.gutenberg.org/} \cite{brooke2015gutentag,bamman2020annotated}. The idiosyncrasies of literary text present several challenges to NLP models for named entity recognition, coreference resolution, character clustering, event detection, and speaker identification. The typical length of a text is several thousands of tokens, and the format and structure of the content vary widely depending on the genre, topic, time-period, and author of the text. Characters are referred to by various aliases, often incorporating notions of familial relations ({\it her father, Mr., Mrs., and Miss Bennet}) or social titles ({\it the baron}); mentions such as the former also can refer to different entities if used by different speakers ({\it my father}).

Consider, for example, the very first quotation in Jane Austen's \textit{Pride and Prejudice}: 
\begin{center}
    \textit{``My dear Mr.\@ Bennet," said his lady to him one day, ``have you heard that Netherfield Park is let at last?"}
\end{center}
Identifying the speaker of this quotation involves making several inferences: that the person being spoken to is Mr.\@ Bennet, that the mention ``his lady" refers to ``Mr.\@ Bennet's" lady, and that this is a proxy for Mr.\@ Bennet's wife, who must be Mrs.\@ Bennet (the first explicit mention of Mrs.\@ Bennet is only in Chapter 2; several other characters are introduced to us in the meantime).

Previous work attempting to solve this task of identifying the speaker of a quotation in the text, or quotation attribution, has explored both rule-based systems \cite{glass2007naive,muzny2017two} and machine learning models that are trained on an annotated dataset of quotations and speakers \cite{elson2010automatic,he2013identification,o2012sequence}. Some of these approaches treat the task as a two-step problem, where quotations are first attached to mentions, and mentions are then attached to a canonical character name. The datasets developed for this task reflect this variation in methodology; some of them are annotated with quotation--mention--speaker information \cite{elson2010automatic,o2012sequence,muzny2017two}, whereas others skip the intermediate mention annotation \cite{he2013identification}. 

In this work, we present a new dataset, the Project Dialogism Novel Corpus (PDNC), comprising 22 full-length novels in which all quotations have been identified and annotated for speaker, addressees (who is being spoken to), characters mentioned, and the referring expression outside the quotation that indicates the speaker (if present). Our contributions are as follows:
\begin{itemize}
    \item PDNC is by an order of magnitude the largest dataset of annotated quotations for literary texts in English, in terms of the number of tokens covered, the number of annotated quotations and characters, as well as the number of character mentions (even though we limit ourselves to mentions within quotations). 
    \item We release, along with the dataset, a comprehensive set of annotation guidelines that cover several idiosyncrasies of literary texts, and which we hope will help standardize future annotation work in this domain.
    \item We evaluate two state-of-the-art quotation attribution systems on this dataset, which obtain average accuracies of 0.62 and 0.63 respectively. We also evaluate a simple semi-supervised classification baseline that achieves competitive results.
    \item We use our annotations to analyze the performance of these models and pinpoint common failure points, which will help inform future work in this area. 
\end{itemize}
All data and code associated with this work will be made publicly available at \url{https://github.com/Priya22/pdnc-lrec2022}.


\section{Background}
We review past datasets of quotations in the literary domain, as well as automatic models for the task of quotation attribution.
\subsection{Prior Datasets}
The Columbia Quoted Speech Attribution (CQSA) corpus from \newcite{elson2010automatic} contains annotations for 3176 instances of quoted speech from 4 novels by each of 4 authors, and 7 short stories from 2 others; only parts of the full-length novels are annotated. Quotations are annotated at the mention-level, i.e, the speaker is chosen from a set of candidate mentions that occur in the nearby context. These mentions are then resolved to speakers by using an off-the-shelf coreference tool. \newcite{he2013identification} annotate a dataset of three novels, \textit{Pride and Prejudice}, \textit{Emma}, and \textit{The Steppe}; the latter two are also present in the CSQA corpus. Their annotation method links quotations directly to canonical characters, rather than mentions. \newcite{muzny2017two} released the QuoteLi dataset, comprising 3103 quotations annotated with both mention and speaker information. The quotations are drawn from the same three novels as those of \newcite{he2013identification}. Finally, \newcite{sims2020measuring} annotate the first 2000 tokens of 100 novels from the LitBank dataset\footnote{https://github.com/dbamman/litbank}. Quotations are linked to a unique speaker from a predefined list of entities. Though this dataset spans the largest number of novels (100), the restricted range of tokens considered results in only 1765 total annotations.

LitBank also contains annotations for coreference, for the same set of 2000 tokens across 100 novels. A total of 29,103 tokens are annotated, of which 24,180 refer to a person, and the rest to other named entities such as places, organizations, vehicles, etc \cite{bamman2020annotated}. Prior to this, \newcite{vala2016annotating} annotated coreference in \textit{Pride and Prejudice}.

\subsection{Models of Quotation Attribution}
\newcite{elson2010automatic} proposed a classification approach for quotation attribution that classifies quotations into one of several types based on whether the speaker is explicitly indicated by an adjoining expression (explicit), appears without an attribution (implicit), is indicated by an anaphoric mention, is part of a dialogue chain, etc (see Table \ref{tab:qtype_ex} for examples of each quotation type). A separate classifier is trained for each of these cases, taking as input a feature vector that encodes information relating to positions of mentions and quotations surrounding the target. Their model achieves an accuracy of 83\% on their dataset, but uses gold labels as part of the pipeline. 

\newcite{o2012sequence} treat the task as a sequence decoding problem, where the set of speaker attributions in a document is treated as a text sequence to be predicted; i.e, the decision for the current quotation is made based on the previous $n$ attribution labels. While this method works well for news data, it fails to beat a rule-based baseline for literary texts. \newcite{he2013identification} approach quotation attribution as a ranking problem between candidate speakers; their SVM-based ranking model selects a speaker based on a feature vector comprising contextual and topic information.

\newcite{muzny2017two} describe a two-step process for quotation attribution, where quotations are first linked to mentions, and mentions to entities. Each step is composed of a set of deterministic sieves, designed to capture cases of increasing complexity. For example, the first sieve looks for explicit trigram patterns of Quote--Speech Verb--Mention. This system is described further in Section \ref{sec-muz}. 

\newcite{Hammond2020TheWT} describe a semi-supervised classification approach to quotation attribution that, similar to those of \newcite{elson2010automatic} and \newcite{he2013identification}, builds a feature vector and trains a classifier to predict the speaker. Their features are based primarily on lexical and syntactic features drawn from work in computational stylometry, and uses an iterative classification approach where high-confidence predictions of the classifier are repeatedly incorporated into the training set.


\begin{table*}[]
    \centering
    \adjustbox{width=\textwidth}{
    \begin{tabular}{p{6cm} l}
    \textbf{Quotation} & \textbf{Annotations} \\ \hline
    
    \textit{``\underline{You} must not be too severe upon \underline{yourself},"} replied  \textbf{Elizabeth} & \specialcell{\textbf{Speaker: }Elizabeth Bennet \\
     \textbf{Addressees: }(Mr. Bennet, Kitty) \\ 
     \textbf{Quote type: }Explicit \\
     \textbf{Referring Expression: } replied Elizabeth \\
     \textbf{Mentions: } (`you', Mr. Bennet), (`yourself', Mr. Bennet)} \\
     \hline
     With an air of indifference \textbf{he} soon afterwards added:
     \textit{``How long did \underline{you} say \underline{he} was at Rosings?"} & \specialcell{\textbf{Speaker: }George Wickham \\
     \textbf{Addressees: }Elizabeth Bennet \\ 
     \textbf{Quote type: }Anaphoric \\
     \textbf{Referring Expression: } he soon afterwards added \\
     \textbf{Mentions: } (`you', Elizabeth Bennet), (`he', Colonel Fitzwilliam)} \\ \hline
     \textit{``But not before \underline{they} went to Brighton?"} & \specialcell{
     \textbf{Speaker: }Elizabeth Bennet \\
     \textbf{Addressees: }Jane Bennet \\ 
     \textbf{Quote type: }Implicit \\
     \textbf{Referring Expression: }  \\
     \textbf{Mentions: } (`they', [George Wickham, Lydia])
     }\\
     \hline
    \end{tabular}}
    \caption{Annotations for three sample quotations from PDNC, one for each quotation type. The speaker in each example is highlighted in bold, and mentions within quotations are underlined.}
    \label{tab:qtype_ex}
\end{table*}

\section{The Project Dialogism Novel Corpus}
We draw our novels from open-source texts available on the Project Gutenberg platform. In selecting these novels, our aim has been to annotate texts in a variety of genres (literary fiction, children’s literature, detective fiction, and science fiction are represented); from the LitBank and QuoteLi corpora, to facilitate comparison and validation; and of broad interest to a variety of scholars while still relevant to our group’s interest in stylistic diversity and dialogism \cite{Hammond2020TheWT,vishnubhotla2019fictional}. Further, we have chosen to annotate multiple novels by Jane Austen, in order to facilitate comparative analysis of a single author’s oeuvre (Austen was chosen because she is included in all existing corpora). 

\subsection{Annotated Attributes}
\label{sec-attr}
Each quotation in our corpus of texts is annotated with the following attributes:
\begin{enumerate}
    \item \textbf{Speaker:} The character uttering the quotation. We limit each quotation to having a single speaker; certain special cases are highlighted in Section~\ref{ann-guid}.
    \item \textbf{Addressee(s):} The set of character(s) being addressed by the speaker. This includes any character that is in the vicinity of the speaker and can ``hear" the uttered quotation.
    \item \textbf{Quotation Type:} Following previous work, we distinguish between explicit, anaphoric, and implicit quotations. See Table \ref{tab:qtype_ex} for an example of each.
    \item \textbf{Referring Expressions:} For explicit and anaphoric quotations, we obtain the part of the text that indicates who the speaker is, the verb for the action of speaking, and sometimes, also the addressees.
    \item \textbf{Mentions:} Finally, we also annotate all characters who are mentioned within a quotation, either explicitly by name or through a pronoun or pronominal phrase. Each mention is linked to the character or set of characters that it refers to.
\end{enumerate}

In addition, each novel is also annotated with a list of characters present in the novel. Each character is associated with a ``main name" (e.g., Elizabeth Bennet), as well as a set of aliases by which they are referred to in the text (e.g., Lizzy, Liz, Elizabeth). The character list includes any character who either speaks, is addressed, or is mentioned in a quotation; therefore we also have characters who are never explicitly assigned a proper name, such as ``The Old Man in the Crowd". 

\subsection{Dataset Statistics}

We list key characteristics of PDNC in Table \ref{tab:pdnc_stats}. A total of 35,978 quotations are identified and annotated for the attributes listed in Section \ref{sec-attr}. On average, we have 1.79 aliases per character, and 1.82 mentions annotated per quotation. Of the 992 characters in our character lists, 655 are speakers of a quotation; of these, 321 characters can be classified as ``minor", having 10 or fewer spoken quotations. Margaret Schlegel from \textit{Howards End} is the most loquacious character across all novels, with 1040 quotations, followed by Jake Barnes from \textit{The Sun Also Rises}, Katherine Hilbery from \textit{Night and Day}, and Anne Shirley from \textit{Anne of Green Gables}.

Figure \ref{fig:qtype_dist} in Appendix \ref{app:qtypes} shows the distribution of quotation types across all novels. We see that implicit quotations make up the largest percentage of annotations ($\sim$37\%), followed by explicit ($\sim$33\%) and anaphoric ($\sim$29\%) quotation types, though the distribution shows a large spread. \textit{Alice in Wonderland} consists mostly of explicit quotations (84\%), whereas Dostoevsky's \textit{The Gambler} is at only 12\%.

We note that PDNC is by far the largest dataset of annotated quotations for works of English Literature. A comparison with previous datasets is presented in Table \ref{tab:prev_data}. Even though we annotate only for mentions within quotations, our count of 62,587 mention annotations is much larger than LitBank's 29,103.

PDNC also contains the largest number of tokens per document (79,745), since we annotate entire novels rather than portions of each. We think that this is an invaluable resource for several open problems in the computational analysis of literature, allowing for tracking character mentions across larger spans of text, studying changes in character style, emotions, and character networks throughout the course of a novel, and the variation of each of these with author and genre.

\begin{table*}[]
    \centering
    \adjustbox{width=\textwidth}{
    \begin{tabular}{l p{3cm} rrrr}
        \textbf{Novel} & \textbf{Author} & \textbf{\# Tokens} & \textbf{\# Quotations} & \textbf{\# Characters} & \textbf{\# Mentions} \\
        \hline
\textit{A Handful Of Dust} & Evelyn Waugh &70299&2617&104&3198\\
\textit{A Room With A View}& E. M. Forster &67434&1989&67&3111\\
\textit{Alice’s Adventures in Wonderland} &Lewis Carroll & 26826&1048&51&683\\
\textit{Anne Of Green Gables} & Lucy Maud Montgomery &103291&1779&114&5168\\
\textit{Daisy Miller} & Henry James &22007&725&10&1021\\
\textit{Emma} & Jane Austen & 161070&2116&18&6318\\
\textit{Howards End} & E. M. Forster&112674&3131&56&4358\\
\textit{Night And Day} &Virginia Woolf &170706&2800&54&3575\\
\textit{Northanger Abbey} & Jane Austen &78081&1017&20&2358\\
\textit{Persuasion} & Jane Austen &83695&702&35&2186\\
\textit{Pride And Prejudice} & Jane Austen &122692&1708&77&4797\\
\textit{Sense And Sensibility} & Jane Austen &120810&1545&25&4676\\
\textit{The Age Of Innocence}& Edith Wharton & 103062&1600&55&2556\\
\textit{The Awakening} & Kate Chopin &50234&738&22&981\\
\textit{The Gambler} & Fyodor Dostoevsky (Trans.\@ C.J. Hogarth) &61508&1068&26&2057\\
\textit{The Invisible Man} & H. G. Wells &49956&1274&33&926\\
\textit{The Man Who Was Thursday} & G. K. Chesterton &58352&1357&31&1700\\
\textit{The Mysterious Affair At Styles} & Agatha Christie &57302&2226&30&3485\\
\textit{The Picture Of Dorian Gray} & Oscar Wilde &80483&1501&45&3336\\
\textit{The Sign of the Four} & Sir Arthur Conan Doyle &43872&891&36&1784\\
\textit{The Sport Of The Gods} & Paul Laurence Dunbar &41470&830&38&1524\\
\textit{The Sun Also Rises} & Ernest Hemingway &68585&3316&45&2789\\
\hline
\textbf{Total}&&1754409&35978&992&62587 \\ \hline
    \end{tabular}}
    \caption{The set of novels annotated in PDNC, with the number of annotated quotations, characters, and mentions in each.}
    \label{tab:pdnc_stats}
\end{table*}

\begin{table*}[]
    \centering
    \begin{tabular}{l rrrrr}
       \textbf{Corpus}  & CQSA (2010)  & \newcite{he2013identification} & \newcite{muzny2017two} & LitBank (2020) & PDNC (2021)\\ \hline
       \textbf{\# Texts} & 6 & 3 & 3 & 100 & 22 \\ 
       \textbf{\# Quotations} & 3176 & 1901 & 3103 & 1765 & 35978  \\
       \hline
    \end{tabular}
    \caption{A comparison of PDNC with previous datasets for quotation attribution in literary texts.}
    \label{tab:prev_data}
\end{table*}


\section{PDNC: The Annotation}
In this section, we describe our annotation process, from developing the guidelines to preprocessing the texts, the annotation platform, and how we resolved disagreements between annotators.

\subsection{Annotation Platform}
We designed our annotation platform from scratch as a web-based interface. A screenshot of the interface is shown in the Appendix, Figure \ref{fig:ann-soft}. The main components include the character list, which allows the annotator to add and remove characters and associated aliases; the text box, which highlights quotations and mentions within the text (different color codes indicate the type and annotation status of the quotation or mention spans); and the annotation area, where values for the desired attributes of a quotation or mention can be set by the annotator. The platform also includes an interface that takes as input two sets of annotations of the same text and generates a file with any disagreements that occur for an annotated attribute, including mis-matches in character lists.

\subsection{Annotation Process}
All our annotators were university-level literature students familiar to one of the authors. Each novel in our corpus was annotated separately by two annotators, and the resulting annotations were then compared to generate a list of ``disagreements". Disagreements were grouped by quotation, and occur when the annotations do not match for any of the attributes listed in Section \ref{sec-attr}. The two annotators then went through a consensus exercise, where they discussed all disagreements, re-annotated the relevant quotations, and once again checked for disagreements (in practice, no more than three rounds of consensus were necessary).

\subsection{Pre-processing the texts}
The raw text for each novel is obtained from the Project Gutenberg platform. This is then processed using the GutenTag software\footnote{https://gutentag.sdsu.edu/} from \newcite{brooke2015gutentag}, which outputs an initial list of characters and aliases, and also identifies quotations within the text. We also pre-identify mentions within each quotation by looking for occurrences of any character names, aliases, or words from a predefined list of pronouns.

\subsection{Annotation Guidelines}
\label{ann-guid}
The complexity of narrative structure and style of literary novels means that several ambiguities can arise while determining any of the annotated attributes. We developed a comprehensive set of guidelines that attempt to cover as many as possible of the cases that we came upon in our texts. These guidelines underwent several revisions as we progressed through different novels, and were informed by feedback from our annotators as well as the authors of this work. We make the complete set of guidelines publicly available and hope it will help guide future work in this area. We highlight a few interesting cases below:

\begin{itemize}
\item Special aliases: Narrators of first-person narratives receive the special alias ``\_narr"; when more than one character speaks a quotation in unison, it is attributed to ``\_group"; when the identity of the speaker is unknowable in context, it is attributed to ``\_unknowable".
\item Multiple addressees: In situations in which many characters are present, our guidelines designate an addressee as anyone ``whom the speaker seems to believe can hear them." 
\item Locating referring expressions: Our guidelines include explicit instructions for annotating referring expressions in cases in which they are difficult to annotate, in which they introduce long or multi-part quotations, and in which multiple referring expressions are applied to single quotation.
\end{itemize}

\section{Quotation Attribution}
We now turn our focus to the analysis of quotation attribution models, where the task is to identify ``who said what". Building an automated attribution system from scratch is generally a multi-step process: we first need to identify quotations in the text, build a list of characters and their aliases, and then attribute each quotation either directly to character, or first to a mention followed by an additional coreference resolution step to identify the associated character.

\subsection{Review of Existing Systems}
We briefly describe two models for quotation attribution that are the current state-of-the-art. 

\subsubsection{A Two-Stage Sieve Approach}
\label{sec-muz}
\newcite{muzny2017two} propose a deterministic, two-step, approach to quotation attribution that relies on several sieves of increasing complexity to first link each quotation to a mention, and then link the mention to a character entity. The latter step involves applying a co-reference resolution model to the text. Since our focus is primarily on the quotation attribution, we briefly describe the main sieves associated with the first step:
\begin{enumerate}
    \item \textbf{Trigram Matching (Tri-1)}: This identifies patterns of the type Quote-Mention-Speech Verb, or Quote-Speech Verb-Mention, to extract quotations where the speaker is indicated by the associated referring expression (e.g., {\it ``she said"}, or {\it ``said Elizabeth"}).
    \item \textbf{Dependency Parses (Dep-2)}: This inspects dependency parses of sentences on either side of the target quotation for speech verbs with an \texttt{nsubj} relation that points to a character mention.
    \item \textbf{Single Mention Detection (Single-3)}: This looks for instances where there is only a single mention in the non-quotation text of the associated paragraph, and attributes the quotation to that mention.
    \item \textbf{Vocative Detection (Voc-4)}: This looks for vocative patterns involving mentions in the previous quotation (e.g., {\it ``are you sure, Lizzy?"}), and links the quotation to the the associated mention.
    \item \textbf{Paragraph Final Mention (Par-5)}: This attributes a quotation occurring at the end of a paragraph to the final mention of the previous sentence.
    \item \textbf{Conversational Pattern (Conv-6)}: This looks for consecutive sequences of quotations (i.e, uninterrupted by non-quote text), and links an unattributed quotation to the speaker of the quotation two steps behind. Muzny et~al.\@ specify a less-restricted version of this where the requirement of ``uninterrupted by non-quote text" is removed.
\end{enumerate}

The sieves, in order, deal with quotations in increasing order of the difficulty of attribution. The easy cases, such as explicit and most anaphoric quotations, are captured by the first two sieves; the latter ones deal with the more complex, implicit quotations that require additional knowledge of the surrounding context. 

\subsubsection{BookNLP}

BookNLP\footnote{https://github.com/booknlp/booknlp} is a tool for natural language processing of literary texts (and other long documents) in English. The pipeline performs, among other things, dependency parsing, named entity recognition, coreference resolution, quotation attribution, and referential gender inference. The latest version of BookNLP is trained on LitBank's annotations of character entities \cite{bamman2020annotated}  and quotations \cite{sims2020measuring}. While the exact model for quotation attribution is not described in a publication, we infer from the code that it uses a BERT-based model that takes as input the quotation text and its surrounding context, and links each quotation to a character mention. Mention-to-entity resolution is performed by a separate pipeline step that precedes quotation attribution. 

\newcite{muzny2017two} specify that they use BookNLP's coreference resolution system for the mention--entity step of their pipeline, though at the time of publication, the latest version of BookNLP was not yet released. Since our focus here is on evaluating models of quotation attribution, separately from coreference resolution, we plug in the latest outputs of BookNLP's coreference resolution system into the two-stage attribution approach of \newcite{muzny2017two}.

\begin{table*}[ht]
    \centering
    \adjustbox{}{
    \begin{tabular}{lrrrr r |rr}
    & \multicolumn{4}{c}{\textbf{State-of-the-art}} && \multicolumn{2}{c}{\textbf{Stylometric}} \\ \hline
        \textbf{Novel} & \textbf{\# Identified} & \textbf{\# Eval} & \textbf{Muzny et al.} & \textbf{BookNLP} && \textbf{\# Eval} & \textbf{Stylo}  \\
        \hline
\textit{A Room With A View}&2071&1857&0.58&\bf 0.59 && 1424 & 0.57\\
\textit{Alice In Wonderland}&1122&965&\bf 0.95&0.93 && 157 & 0.76\\
\textit{Anne Of Green Gables}&1841&1726& \bf 0.88&0.86 && 660 & 0.60\\
\textit{Daisy Miller}&749&713&0.70& \bf 0.74 && 390 & 0.73\\
\textit{Emma}&2108&1935&0.61& \bf 0.62 && 1422 & 0.60\\
\textit{Handful Of Dust}&2732&2502&0.58& \bf 0.59 && 1956 & 0.54\\
\textit{Howards End}&3304&2917&0.61& \bf 0.66 && 2195 & 0.56\\
\textit{Night And Day}&2901&2619& \bf 0.74&0.72 && 1776 & 0.68\\
\textit{Northanger Abbey}&1072&1001&\bf 0.59&0.54 && 734 & 0.66\\
\textit{Persuasion}&786&655&\bf 0.69&0.63 && 330 & 0.33\\
\textit{Pride And Prejudice}&1779&1681&0.63& \bf 0.64 && 1133 & 0.48\\
\textit{Sense And Sensibility}&1546&1472&0.63& \bf 0.64 && 886 & 0.31\\
\textit{The Age Of Innocence}&1912&1466&0.44& \bf 0.45 && 1235 & 0.75\\
\textit{The Awakening}&782&705&0.59& \bf 0.62 && 517 & 0.65\\
\textit{The Gambler}&1128&1012&0.40& \bf 0.42 && 920 & 0.74\\
\textit{The Invisible Man}&1277&1103&\bf 0.80&0.79 && 585 & 0.52\\
\textit{The Man Who Was Thursday}&1339&1264& \bf 0.78&0.76 && 479 & 0.44\\
\textit{The Mysterious Affair At Styles}&2228&2103& \bf 0.50&0.42 && 1791 & 0.66\\
\textit{The Picture Of Dorian Gray}&1539&1450&0.59& \bf 0.66 && 1068 & 0.63\\
\textit{The Sign Of the Four}&900&815&0.42& \bf 0.44 && 710 & 0.72\\
\textit{The Sport Of The Gods}&885&783&0.46& \bf 0.50 && 625 & 0.44\\
\textit{The Sun Also Rises}&3324&3219&0.52& \bf 0.55 && 2223 & 0.65\\
\hline
\textbf{Total} & 37325 & 33963 & 0.62 & \bf 0.63 && 23216 & 0.59\\ \hline
    \end{tabular}}
    \caption{Accuracy scores for the quotation attribution systems from Muzny et al.\@, BookNLP, and the stylometric classifier (Stylo). The first numerical column in each row for the SoTA models is the number of quotations identified by BookNLP, the second is the number of quotations for which the predicted cluster of speaker mentions could be matched with our annotated list of characters. For the Stylo model, \# Eval is the number of non-explicit quotations by major characters in PDNC for that novel.}
    \label{tab:sota_perf}
\end{table*}

\subsubsection{A Semi-Supervised Stylometric Approach}
\label{sec-stylo}
One of the key uses of our corpus is in work on dialogism, i.e. variation in the speaking styles of characters in a novel as compared to one another and to the narrator. \newcite{Hammond2020TheWT} propose a stylometric, semi-supervised classification approach to quotation attribution that relies on the stylistic characteristics of the quotation text to identify the speaker. We test here a slightly modified version of that approach, the details of which are in Appendix \ref{app-stylo}. 

Briefly, for each quotation in our dataset, we extract a set of features based on the annotated attributes: the text of the quotation, the referring expression (if present), and the set of mentions. These features are drawn from prior work in computational stylometry \cite{altakrori2021topic,vishnubhotla2019fictional}. The feature vectors are passed to a classifier that is trained to predict the the speaker in an $n$-way classification setup. The model follows a semi-supervised approach that iteratively extracts high-confidence predictions from the test set and adds them to the training set for the next round of classification. 

Note that this model does not function as a stand-alone quotation attribution system, since it assumes access to both a fixed list of characters and gold speaker labels. We merely test it on the PDNC dataset to examine the viability of a stylometric approach to the speaker attribution problem, and as a complement to existing approaches.

\begin{table*}[h]
\centering
    \begin{tabular}{lrr rr rr}
    & \multicolumn{2}{c}{\textbf{Explicit}} & \multicolumn{2}{c}{\textbf{Anaphoric}} & \multicolumn{2}{c}{\textbf{Implicit}} \\ \hline
    \textbf{Method} & \textbf{\# Qs} & \textbf{Acc.} & \textbf{\# Qs} & \textbf{Acc.} & \textbf{\# Qs} & \textbf{Acc.}  \\ \hline
Muzny et al. & 11545&0.96&9855&0.48&12551&0.41\\
BookNLP & 11545&0.94&9855&0.46&12551&0.46\\
Stylo & 11556& -- &10072&0.67&13133&0.54 \\ 
\hline
    \end{tabular}
    \caption{Breakdown of the performance of our models by quotation type. Stylo refers to the stylometric model.}
    \label{tab:qtype-perf}
\end{table*}

\subsection{Experimental Setup}
To test the \newcite{muzny2017two} model, we use the Python re-implementation from \cite{sims2019literary}, as it fits well with our Python pipeline (the original implementation is in Java, and integrated with the StanfordCoreNLP pipeline). We use the latest version of BookNLP to identify quotations within the text and a list of character clusters. The latter is obtained as an output of the coreference resolution module, which clusters together mentions within the text that are presumed to refer to the same entity. As such, the model does not build a list of canonical character names with which to associate quotations; rather, each quotation is attributed to an entity cluster. 

This presents a slight problem for evaluating their performance based on our gold standard annotations. Consider a character cluster identified by BookNLP as follows:
\textit{\{my, her, mingott, i\}}. This is identified as the speaker of a quotation, whose speaker label in PDNC is \textit{Mrs.\@ Lovell Mingott}. However, we also have another character in the same novel, \textit{Uncle Lovell Mingott}. The ambiguity in matching characters can in this particular case be resolved by inferring the gender of both characters via the associated pronouns; however, we observed quite a few cases where either a resolution was not possible (e.g., the name \textit{Mingott} can refer to any member of the Mingott family; even \textit{Miss Mingott} could refer to more than one unmarried female of the Mingott family, depending on the context), or the character cluster contained conflicting pronouns (both \textit{her} and \textit{him} appeared along with the name \textit{Mingott}).

In our evaluation of the systems, we do not consider the cases where this ambiguity could not be resolved; this results in the evaluation size being smaller than the set of identified quotations. We report the difference in these sizes in Table \ref{tab:sota_perf}; for most novels, this number lies in the lower hundreds.

For the stylometric classification model, we limit ourselves to characters with at least 10 annotated quotations, in order to avoid the long tail of minor characters.
To further mitigate the class imbalance issue, we oversample from the minority classes. 
We use a Logistic Regression classifier with a grid search over the regularization hyperparameter.
The initial training and test sets for each novel are based on quotation types: explicit quotations are assigned to the training set, and the rest form the test set.
The probability threshold for each iteration of the classification is dynamically determined as a mean of the probabilities over correct predictions; we observed that this parameter varies from novel to novel, and that the classification setup is quite sensitive to this value. 

\begin{table*}[h]
    \centering
    \begin{tabular}{p{2cm} p{12cm}}
    \textbf{Sieve} $\;\;\;\;$& \textbf{Example} \\ \hline
    Tri-1 & ``Gad," Archer heard Lawrence Lefferts say, ``\textit{not one of the lot holds the bow as she does}" and Beaufort retorted ``Yes but that's the only kind of target she'll ever hit." \textbf{Speaker:} Lawrence Lefferts  $\;\;\;\;$ \textbf{Predicted: } Julius Beaufort\\ \hline 
    Dep-2 & Mr. Welland, beaming across a breakfast table miraculously supplied with the most varied delicacies, was presently saying to Archer \textit{``You see, my dear fellow, we camp we literally camp. I tell my wife and May that I want to teach them how to rough it."} \textbf{Speaker:} Mr Welland $\;\;\;\;$ \textbf{Predicted: }Newland Archer \\ \hline
    Voc-4 & Q1 (Newland): ``Your mother?" \\
    & Q2: \textit{``Yes  the day before she died."}  \\ & \textbf{Speaker:} Dallas Archer  $\;\;\;\;$ \textbf{Predicted:} your mother \\ \hline

    \end{tabular}
    \caption{Example quotations from \textit{The Age of Innocence} that are mis-attributed by sieves of Muzny et al.'s attribution model. The quotation under consideration is italicised.}
    \label{tab:muz_fail}
\end{table*}

\begin{table}[h]
\adjustbox{width=0.5\textwidth}{%
    \begin{tabular}{l rr rr rr}
    & \multicolumn{2}{c}{\textbf{Explicit}} & \multicolumn{2}{c}{\textbf{Anaphoric}} & \multicolumn{2}{c}{\textbf{Implicit}} \\ \hline
    \textbf{Sieve} & \textbf{\# Qs} & \textbf{Acc.} & \textbf{\# Qs} & \textbf{Acc.} & \textbf{\# Qs} & \textbf{Acc.}  \\ \hline
Tri-1 &6500&0.98&5750&0.50&24&0.42\\
Dep-2 &3529&0.96&2872&0.46&271&0.43\\
Single-2 &1212&0.82&831&0.26&993&0.43\\
Voc-4&55&0.29&68&0.22&1669&0.46\\
ParFinal-5&0&--&1&1.00&6&0.33\\
ConvPat-6&156&0.24&207&0.27&7208&0.52\\
BASE-7&93&0.26&126&0.29&2380&0.30\\ \hline
    \end{tabular}}
    \caption{Breakdown of the performance of each sieve from \protect \newcite{muzny2017two} by quote type. \textbf{\# Qs} indicates number of quotations.}
    \label{tab:muz-perf}

\end{table}

\subsection{Results}
Table \ref{tab:sota_perf} shows the performance of the two state-of-the-art (SoTA) attribution systems and the stylometric classifier on the PDNC novels. Note that, for the former, since we use BookNLP as a common pipeline for the quotation identification and character name clustering steps, both systems are evaluated on the same set of quotations. For the stylometric model, the accuracy is calculated on the set of non-explicit quotations by non-minor characters (at least 10 annotated quotations).

We see that there is a large variation in the performance across novels, for all models. Certain novels, such as \textit{Anne of Green Gables}, seem easier to attribute for both of our SoTA models; likewise, others seem to present difficulties across the board (\textit{The Sport of the Gods}). \textit{Alice in Wonderland}, in particular, achieves near-perfect accuracy scores. This can partly be attributed to the fact that Alice is by far the most common speaker, contributing to nearly 42\% of all quotations in the novel, and nearly 84\% of the quotations are explicit.

With the stylometric model, which is evaluated only on non-explicit quotations, we do quite well on certain novels that the SoTA systems struggle with. For \textit{The Age of Innocence}, for example, the stylometric model correctly attributes 75\% of the implicit and anaphoric quotations, which account for nearly 80\% of the total annotated quotations. By contrast, both the Muzny and BookNLP models achieve accuracies of about 45\%.







\subsubsection{Performance by Quotation Type}
Table \ref{tab:qtype-perf} presents a breakdown of the performance of each of our three models by quotation type. Note that since we use explicit quotations as the training set for the stylometric system, we do not report an accuracy score in that cell. Both the Muzny and BookNLP models perform quite well on explicit quotations. As expected, implicit quotations are the hardest to attribute. That anaphoric quotations do not fare much better indicates that the coreference resolution part of the attribution pipeline is responsible for many mis-attributions; we verify this hypothesis in the next section. 

\subsection{Evaluating the Sieves}
We examine how often the heuristic sieves proposed by \newcite{muzny2017two} hold up across all our novels. For each sieve, we try to answer questions with regard to the number of quotations of each type captured by the sieve, it's performance on these quotations, and the possible reasons for mis-attributions.

We first take a qualitative look by examining the performance of the model for one of the novels in our corpus, \textit{The Age of Innocence}. 
Table \ref{tab:muz_fail} lists examples quotations from the text that are wrongly attributed based on mentions in the context surrounding the target quotation; many of these occur due to sentence structures that are not straightforward. We observe several such instances in this text and others, indicating that the surrounding contextual information alone may not always be sufficient to attribute quotations. 

However, the most common source of attribution errors that we observed in our analyses was failure of the coreference resolution module. Even with BookNLP's state-of-the-art model, a large number of character clusters either are not associated with a character entity explicitly by name (e.g., \{\textit{you, yourself, your, i, she, her\}} forms one of the clusters), or mix together mentions of several different characters into a single cluster, sometimes with opposing gendered pronouns (\{\textit{herself, my, yourself, archer, his\})}.

\subsubsection{Quantitative Analysis}
Table \ref{tab:muz-perf} details the performance of each of the 6 sieves, along with an additional baseline sieve, BASE-7 (attribute to the most common mention in a 5000-word window surrounding the target quotation), when divided by quotation type. Surprisingly, not all explicit quotations are captured by the trigram and dependency parse sieves, indicating the prevalence of more-complex referring expressions, even with explicit character mentions. We also note that the Paragraph Final Mention sieve rarely comes into play. The accuracy on anaphoric and implicit quotations doesn't exceed 50\% across the board, highlighting again the key role played by the coreference resolution module. 

\subsection{Discussion}
Our results demonstrate the challenges posed by literary novels for quotation attribution. Accuracy scores vary widely across novels for all three models that we evaluate. Implicit and explicit quotations in particular are hampered by the mention-to-entity step of the pipeline, due to the much harder task of coreference resolution in this domain. The stylometric model, which directly predicts speaker labels, does relatively better on these subsets. Though most recent work in this area has moved away from building canonical character lists, instead defaulting to mention clusters, we think that the former approach is better for a standardized evaluation of the task. It is also beneficial for downstream applications that use these outputs, such as analyzing stylistic patterns of individual characters, building networks of speaker interactions, and analyzing broader trends in these across authors and genres.

\section{Conclusion}
We presented a new dataset of quotation annotations for English literary texts, with 35,978 quotations across 22 full-length novels annotated for speaker, addressees, quotation type, referring expression, and mentions. This is the largest dataset of quotations and mentions in this domain. We hope that the comprehensive set of annotation guidelines developed as part of the annotation process will be useful for any future work in this area. We hope to expand PDNC with a more diverse set of texts in the future.

We demonstrated that existing quotation attribution models still have a long way to go in reliably identifying the speaker of a quotation, despite being trained on literary datasets. PDNC provides a new source of training data for these models, and its annotated attributes are also useful in identifying the causes of errors in attribution. We showed that a stylometric classification model serves as a competitive baseline for the task, and would be a useful complement to attribution models.

\section*{Acknowledgements}
We are grateful to the annotators involved in the creation of this dataset, 
Alanna Carolan, Sofia Chabchoub, Leah Duarte, Bisman Kaur, Sol Kim, Sanghoon Oh, Jovana Pajovic, and Beck Siegal.  The first author is supported by funding from the Natural Sciences and Engineering Research Council of Canada, and resources provided by the Vector Institute of Artificial Intelligence. This research was supported by a Connaught New Researcher Award from the University of Toronto.

\section{Bibliographical References}\label{reference}
\bibliographystyle{lrec2022-bib}
\bibliography{lrec2022-pdnc}


\appendix
    \begin{center}
      {\bf \large Appendix}
    \end{center}

\section{Types of Quotations in PDNC}
\label{app:qtypes}
\begin{figure}[h!]
    \centering
    \includegraphics[width=0.4\textwidth]{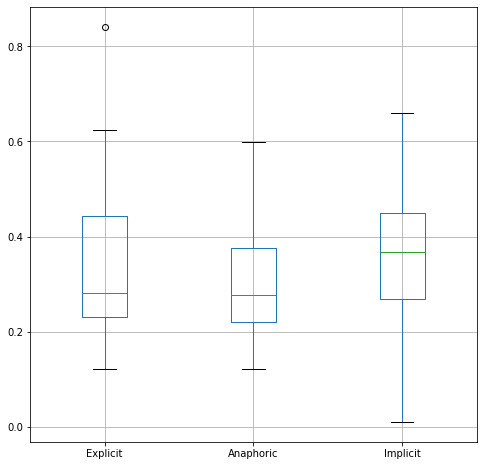}
    \caption{Distribution of quotation types across novels in PDNC. }
    \label{fig:qtype_dist}
\end{figure}
Figure \ref{fig:qtype_dist} shows  box-and-whisker plot of the three quotation types annotated in our dataset. The central region (the box) indicates the ``middle portion" of the data distribution, i.e, the range covered between the first quartile (the 25\% mark) and the third quartile (the 75\% mark), with the median (50\% mark) lying at line inside the box. The whiskers, the dashes on either end of the plot, are at a distance of 1.5 times the inter-quartile length (inter-quartile length is the distance between the first and third quartiles). Points beyond the whiskers are considered outliers.

\begin{figure*}[t!]
    \centering
    \includegraphics[width=0.8\textwidth, height=7cm]{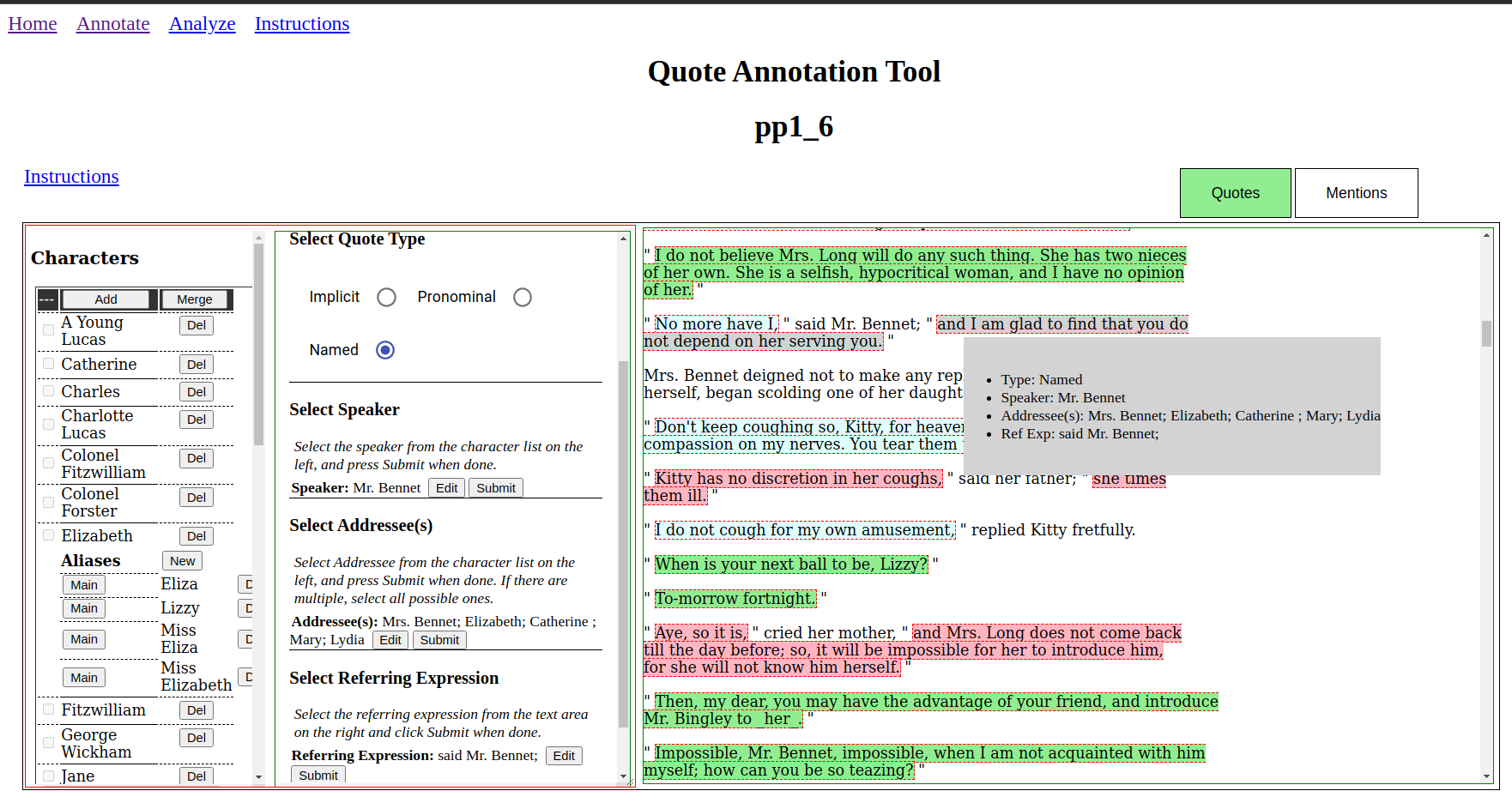}
    \caption{A screenshot of our annotation platform. The difference colors indicate the type of quotation.}
    \label{fig:ann-soft}
\end{figure*}

\section{Stylometric Classification for Quotation Attribution}
\label{app-stylo}
\begin{table*}[htbp]
\centering
\begin{tabular}{p{7cm}}
\hline
\textbf{Lexical Features --- Character-Level}  \\\hline
1. Characters count (N) \\
2. Ratio of digits to N \\
3. Ratio of letters to N \\
4. Ratio of uppercase letters to N \\
5. Ratio of tabs to N \\
6. Frequency of each alphabet (A-Z), ignoring case (26 features) \\
7. Frequency of special characters: \textless\textgreater\%\textbar\{\} []/$\backslash$@\#\~\ +-*=\$\^\ \&\_()' (24 features). \\
\end{tabular}
\begin{tabular}{|p{8cm}}
\hline
\textbf{Lexical Features --- Word-Level}\\\hline
1. Tokens count (T)\\
2. Average sentence length (in characters)\\
3. Average word length (in characters)\\
4. Ratio of alphabets to N\\
5. Ratio of short words to T (a short word has a length of 3 characters or less)\\
6. Ratio of words length to T. Example: 20\% of the words are 7 characters long. (20 features)\\
7. Ratio of word types (the vocabulary set) to T\\
\end{tabular}\\
\begin{tabular}{p{15.5cm}}
\hline
\textbf{Syntactic Features}  \\\hline
1. Frequency of Punctuation: , . ? ! : ; ' " (8 features) \\
2. Frequency of function words from \newcite{OShea.J:2013} (277 features)\\
\hline
\end{tabular}
\caption{List of stylometric features from \protect \newcite{altakrori2021topic}}
\label{tbl:features}
\end{table*}

Here, we describe the classification-based approach to quotation attribution adapted from \newcite{Hammond2020TheWT}.

\subsubsection*{Feature Extraction}
The feature vector is composed of the following sets of features:
\begin{enumerate}
    \item \textbf{Stylometric features: } From the quotation text, we extract a set of 371 features that capture character and word-level lexical and syntactic features of the text. These features were drawn from prior work in authorship attribution and computational stylometry, particularly that of \newcite{altakrori2021topic}. The list of features is in Table \ref{tbl:features}.
    \item \textbf{TF-IDF Counts: } We vectorize the quotation text (excluding stop words used in Feature Set 1 above), the text of the referring expression, and the set of entities mentioned within the quotations, where available, using TF-IDF counts.
    \item \textbf{Lexicon-based features: } For each quotation text, we find the average value of the words in the quotation along a set of lexical dimensions, where the values are obtained via lexicons. The first set of features are the six dimensions of style as described by \newcite{brooke2013multi} --- literary, abstract, objective, colloquial, concrete, subjective, polarity --- with the associated lexicon provided by the authors. The second set of features comes from the NRC Emotion Intensity Lexicon, which associates each word with a real-valued score along eight basic emotions --- anger, anticipation, fear, joy, sadness, and trust --- and two sentiments, positive and negative \cite{Mohammad13}. Finally, we compute these features for the three emotion dimensions of valence, arousal, and dominance, the lexicons for which are obtained from the work of \newcite{vad-acl2018}. 
\end{enumerate}

\subsubsection*{Classification}
Our classification model is a semi-supervised approach that iteratively extracts high-confidence predictions from the test set and adds them to the training set for the next round of classification. Let us assume a dataset of quotation--speaker pairs $(X, y)$, and an initial train and test set of quotation--speaker pairs $(X_{\textit{train\_init}}, y_{\textit{train\_init}})$ and $(X_{\textit{test\_init}}, y_{\textit{test\_init}})$. The classification pipeline proceeds as follows:
\begin{enumerate}
    \item Set $(X_{\textit{train}}, y_{\textit{train}}) \leftarrow (X_{\textit{train\_init}}, y_{\textit{train\_init}})$ and $(X_{\textit{test}}, y_{\textit{test}}) \leftarrow (X_{\textit{test\_init}}, y_{\textit{test\_init}})$.
    \item Extract feature vectors for $X_{\textit{train}}$ and $X_{\textit{test}}$.
    \item Train a classifier $\mathit{Clf}$ on the training data to predict the speaker, $y$.
    \item Obtain the predictions of $\mathit{Clf}$ on the test set, $y_{\textit{pred}}$, and the associated prediction probabilities, $y_{\textit{pred\_probs}}$.
    \item Extract the test instances $(X_{\textit{cand}}, y_{\textit{cand}}) \subseteq (X_{\textit{test}}, y_{\textit{pred}})$ that have a prediction probability greater than some threshold, $y_{\textit{pred\_probs}} \geq T$. 
    \item Add these to the initial train set to obtain the train and test sets for the next round $(X_{\textit{train}}, y_{\textit{train}}) \leftarrow (X_{\textit{cand}}, y_{\textit{cand}}) \cup (X_{\textit{train\_init}}, y_{\textit{train\_init}})$; $(X_{\textit{test}}, y_{\textit{test}}) \leftarrow (X, y) \setminus (X_{\textit{train}}, y_{\textit{train}})$.
    \item Repeat the process from Step 2; break when there is no improvement in test performance for three consecutive iterations, or we hit 20 iterations.
\end{enumerate}
Test instances that have been added to the initial train set in one round can be removed in a subsequent round if they do not satisfy the probability threshold. 
\end{document}